\documentclass[conference]{IEEEtran}
\IEEEoverridecommandlockouts
\usepackage{cite}
\usepackage{amsmath,amssymb,amsfonts}
\usepackage{algorithmic}
\usepackage{graphicx}
\usepackage{textcomp}
\usepackage{xcolor}
\usepackage{booktabs}
\usepackage{multirow}
\usepackage{url}
\usepackage[hidelinks]{hyperref}
\def\BibTeX{{\rm B\kern-.05em{\sc i\kern-.025em b}\kern-.08em
    T\kern-.1667em\lower.7ex\hbox{E}\kern-.125emX}}
\begin{document}

\title{Investigating the Role of Reasoning Augmentation in Parameter-Efficient Fine-Tuning for Implicit Sentiment Analysis}


\author{
\IEEEauthorblockN{1\textsuperscript{st} Hoang Nguyen Minh}
\IEEEauthorblockA{
\textit{University of Information Technology}\\
\textit{Vietnam National University} \\
Ho Chi Minh City, Vietnam \\
23520530@gm.uit.edu.vn}
\and
\IEEEauthorblockN{2\textsuperscript{nd} Huy Luu Gia}
\IEEEauthorblockA{
\textit{University of Information Technology}\\
\textit{Vietnam National University} \\
Ho Chi Minh City, Vietnam \\
23520618@gm.uit.edu.vn}
\and
\IEEEauthorblockN{3\textsuperscript{rd} Dang Van Thin}
\IEEEauthorblockA{
\textit{University of Information Technology}\\
\textit{Vietnam National University} \\
Ho Chi Minh City, Vietnam \\
thindv@uit.edu.vn}
}
\maketitle

\begin{abstract}
Implicit Sentiment Analysis (ISA) demands complex reasoning often absent in standard supervised models. This paper explores the potential of parameter-efficient fine-tuning (PEFT) for ISA using Gemma-7B with Low-Rank Adaptation (LoRA). We introduce a reasoning-aware framework that augments training with structured linguistic rationales distilled from GPT-4o. Experiments on SemEval-2014 benchmarks demonstrate that our framework significantly outperforms BERT-based and prompt-based baselines, achieving parity with 11B-parameter models. Crucially, we identify a domain dependency: while reasoning augmentation enhances performance in subjective contexts (Restaurant) by bridging semantic gaps, it can introduce detrimental noise in objective technical domains (Laptop) through the over-interpretation of factual specifications. To facilitate reproducibility, our code and data are released at the GitHub\footnote{\url{https://github.com/UIT23520530/atac2026}}.
\end{abstract}


\section{Introduction}
\label{sec:introduction}
Sentiment Analysis (SA) is a fundamental task in Natural Language Processing (NLP) aimed at classifying the polarity of text to determine the attitude of the speaker toward specific entities \cite{liu2012sentiment}. While explicit SA has been extensively studied, Implicit Sentiment Analysis (ISA) represents a more intricate frontier, particularly within the scope of Aspect-Based Sentiment Analysis (ABSA). Unlike explicit SA, where sentiment is conveyed through distinct opinion words, ISA relies on semantic context, tone, metaphor, and pragmatic inference \cite{russo-etal-2015-semeval}. The exponential growth of user-generated content on social media—where sentiment is often veiled in sarcasm or cultural references—underscores the critical urgency of advancing ISA research.

Recently, Large Language Models (LLMs) have revolutionized NLP with their vast knowledge bases and superior reasoning abilities. Prior works \cite{zhang-etal-2024-sentiment, xu-etal-2024-improving, kardkovacs-etal-2025-btc} have empirically validated the efficacy of LLMs in general SA, identifying them as powerful tools for opinion mining. However, the intersection of LLMs and ISA presents a distinct challenge. Despite these advancements, ISA research remains predominantly anchored in encoder-only models like BERT or encoder-decoder architectures like T5 \cite{10584152, 10869316, 11214474}. While recent studies have begun to apply LLMs to ISA, they are largely confined to inference-time prompting techniques \cite{fei-etal-2023-reasoning, 10549245, ren2025capital}. Crucially, the potential of fine-tuning LLMs—specifically instruction tuning integrated with Chain-of-Thought (CoT) reasoning traces—remains significantly underexplored in the context of ISA. Given that classifying implicit sentiment necessitates complex, multi-hop reasoning to bridge the gap between surface text and underlying intent, standard fine-tuning may be insufficient. Yet, it remains an open question whether explicitly embedding these reasoning traces into the parameters of the model enhances performance or introduces unnecessary complexity in smaller LLMs.

To address this, we present a comprehensive study with the following contributions:
\begin{itemize} 
   \item We conduct the first systematic evaluation of Gemma-7B for ISA, contrasting standard versus reasoning-aware fine-tuning on the SemEval-2014 benchmarks \cite{li-etal-2021-learning-implicit}. 
   \item We establish a unified, reproducible prompting baseline (encompassing Zero-shot, Few-shot, and CoT) to address inconsistencies in prior comparisons and ensure rigorous benchmarking.
   \item Our analysis reveals a critical domain dependency: while reasoning augmentation enhances performance in subjective contexts, we identify specific linguistic phenomena in objective domains where it introduces noise, offering actionable insights for future research.
\end{itemize}

\section{Related Works}
\label{sec:related_works}

\paragraph{Sentiment Analysis (SA)} The landscape of Sentiment Analysis has been fundamentally reshaped by the advent of LLMs. A comprehensive empirical evaluation by \cite{zhang-etal-2024-sentiment} highlighted the versatility of LLMs across a spectrum of tasks, ranging from standard polarity classification to more granular aspect-based and multifaceted extraction. To address data scarcity in low-resource scenarios, recent studies have increasingly focused on cross-lingual transfer mechanisms, effectively utilizing knowledge sharing to improve model generalization \cite{smid-etal-2025-laca}. Concurrently, researchers have explored instruction-tuning paradigms, integrating in-context learning and advanced reasoning strategies to better align LLM outputs with the nuances of human sentiment inference \cite{xu-etal-2024-improving, shen-etal-2025-zero}.

\paragraph{Implicit Sentiment Analysis (ISA)} Distinguished by the absence of explicit opinion words, ISA poses significantly greater interpretative challenges than its explicit counterpart. Seminal efforts to establish benchmarks include the creation of aspect-level datasets for the English laptop and restaurant domains \cite{li-etal-2021-learning-implicit}, as well as a sentence-level corpus for Vietnamese technology reviews \cite{luu-thin-2025-exploring}. Contemporary methodologies have largely diverged into two streams: (1) Fine-tuned supervised models, predominantly leveraging BERT or T5-style architectures \cite{10584152, 10869316, 11214474}, and (2) Prompting-based approaches that utilize multi-hop reasoning to exploit the inferential capacity of LLMs \cite{fei-etal-2023-reasoning, 10549245, ren2025capital}. Consequently, a critical gap exists for a unified framework that combines the stability of supervised fine-tuning with the reasoning capabilities of LLMs specifically tailored for implicit sentiment inference.

\section{Methodology}
\label{sec:methodology}
\begin{figure*}[t]
    \centering
    \includegraphics[width=\textwidth]{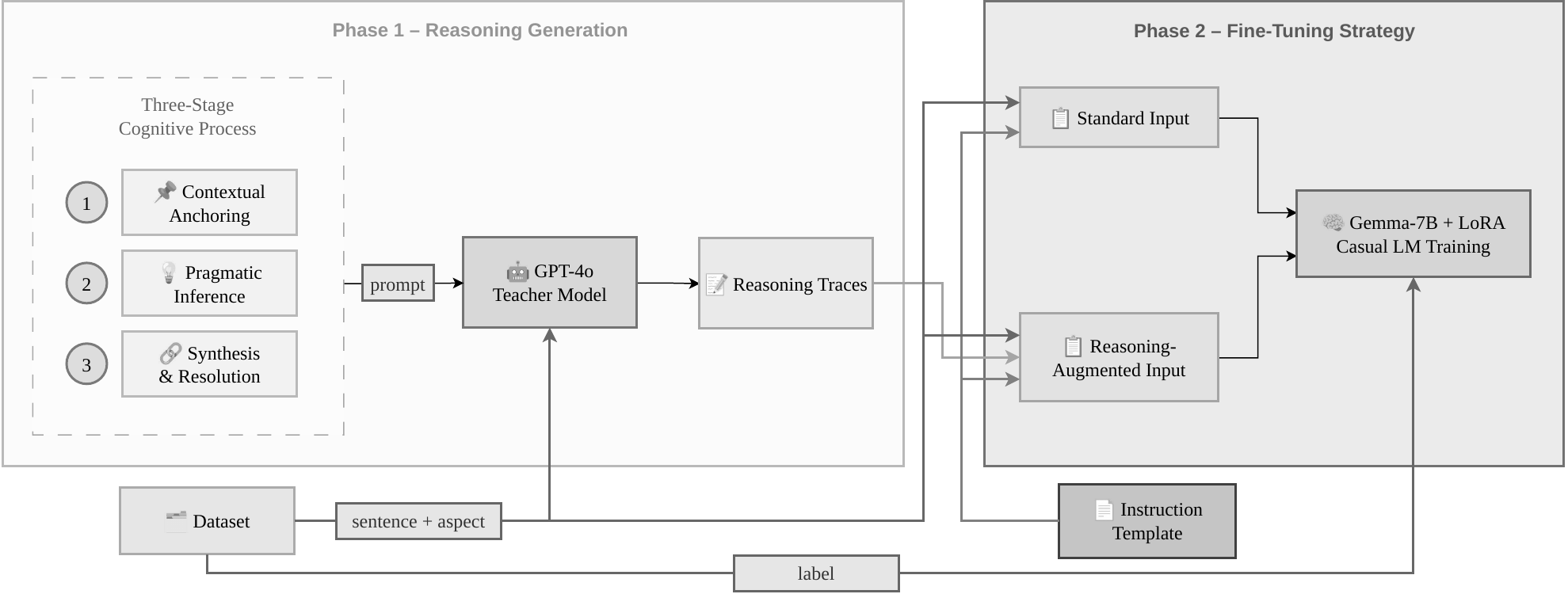}
    \caption{Overview of the proposed two-phase framework, where reasoning traces generated by a teacher model are used to construct reasoning-augmented inputs for LoRA-based fine-tuning of the Gemma-7B model.}
    \label{fig:method}
\end{figure*}

In this section, we formally define the task of aspect-based ISA and present the methodology employed to explore the impact of reasoning augmentation on LLMs. We specifically focus on the Parameter-Efficient Fine-Tuning (PEFT) approach using Low-Rank Adaptation (LoRA) applied to Gemma-7B model. Figure~\ref{fig:method} illustrates the overall framework of our approach, consisting of a reasoning generation phase using a teacher model and a subsequent LoRA-based fine-tuning phase that leverages reasoning-augmented inputs.

\subsection{Problem Formulation}
Let $\mathcal{D} = \{(x_i, a_i, y_i)\}_{i=1}^{N}$ be a dataset consisting of $N$ samples. For each sample, $x_i$ denotes the input sentence (context), $a_i$ represents the target aspect term appearing in the sentence, and $y_i \in \mathcal{Y}$ is the sentiment label associated with the aspect $a_i$, where $\mathcal{Y} = \{\text{Positive, Negative, Neutral}\}$.

Unlike explicit sentiment analysis where opinion words are directly present (e.g., ``good'', ``terrible''), ISA deals with sentences where the sentiment is inferred from factual context or semantic nuance. The goal is to learn a mapping function $f: (x, a) \to y$ that maximizes the probability of predicting the correct sentiment label given the context and aspect.

\subsection{Preliminaries}

\subsubsection{Instruction Tuning}
Instruction tuning adapts a pre-trained LLM to follow natural language instructions. In this work, we formulate the aspect-based ISA task as a generative problem. Given an input sentence $x$ and an aspect $a$, we construct an instruction template $\mathcal{I}$ that guides the model to generate the textual representation of the label $y$.

\subsubsection{Low-Rank Adaptation (LoRA)}
Fine-tuning full parameters of LLMs is computationally expensive. We employ LoRA \cite{hu2022lora}, a parameter-efficient technique that freezes the pre-trained model weights and injects trainable rank decomposition matrices into each layer of the Transformer architecture.

Formally, for a pre-trained weight matrix $W_0 \in \mathbb{R}^{d \times k}$, LoRA constrains the weight update $\Delta W$ by representing it as the product of two low-rank matrices $A \in \mathbb{R}^{d \times r}$ and $B \in \mathbb{R}^{r \times k}$, where the rank $r \ll \min(d, k)$. The forward pass is modified as follows:
\begin{equation}
    h = W_0 x + \Delta W x = W_0 x + ABx
\end{equation}
During training, $W_0$ is frozen, while $A$ and $B$ are optimized. This significantly reduces the number of trainable parameters while maintaining model performance.

\subsection{Reasoning-Augmented Fine-Tuning Framework}
To investigate whether explicit reasoning traces can enhance the ability of the model to detect implicit sentiment, we design a two-phase framework consisting of: (1) Reasoning Generation and (2) Fine-Tuning Strategy.

\subsubsection{Phase 1 – Reasoning Generation (Offline)}
Since standard datasets typically provide only ground truth labels without explanations, we utilize a strong teacher model, GPT-4o, to distill reasoning capabilities. For each training sample $(x_i, a_i)$, we prompt the teacher model to generate a reasoning trace $r_i$ that explains why the aspect $a_i$ in sentence $x_i$ bears a specific sentiment.
\begin{equation}
    r_i \leftarrow \text{GPT-4o}(x_i, a_i, \text{Prompt}_{\text{gen}})
\end{equation}
The generated reasoning $r_i$ acts as an intermediate rationale, bridging the gap between the implicit context and the final sentiment label.

To ensure high-quality rationale generation, we design $\text{Prompt}_{\text{gen}}$ based on a structured linguistic analysis protocol rather than simple open-ended generation. This protocol enforces the teacher model to simulate an expert linguistic analyst through a three-stage cognitive process:
\begin{itemize}
    \item Contextual Anchoring: The model first locates the aspect within the domain context (e.g., distinguishing between a factual description of ``weight'' in a laptop versus a restaurant context) to establish baseline expectations.
    \item Pragmatic Inference: A critical step for implicit sentiment, where the model applies \textit{benefit/harm logic} and evaluates \textit{normative deviation}. The model analyzes whether a factual description (e.g., ``high price'' or ``heavy casing'') represents a deviation from standard norms and assesses the resulting utility or inconvenience to the user.
    \item Synthesis and Resolution: Finally, the model synthesizes explicit signals and implicit pragmatic cues to resolve potential conflicts between surface tone and underlying meaning, producing a cohesive dense paragraph that naturally implies the sentiment polarity.
\end{itemize}

\subsubsection{Phase 2 – Fine-Tuning Strategy}
In this phase, we integrate the distilled reasoning into the training process. To rigorously evaluate the efficacy of our proposed integration, we formulate two distinct input construction strategies for comparison:

\paragraph{Standard Input} First, we define a standard prompt structure that simply concatenates the instruction $\mathcal{I}$, the input sentence $x$, and the aspect $a$. This serves as the baseline to measure performance without reasoning augmentation. The standard input sequence $S_{\text{std}}$ is defined as:
\begin{equation}
    S_{\text{std}} = [\mathcal{I}; x; a]
\end{equation}

\paragraph{Reasoning-Augmented Input} To incorporate the generated rationale, we augment the input space by embedding the reasoning trace $r$ directly into the prompt context. The reasoning-enhanced input sequence $S_{\text{reas}}$ is formulated as:
\begin{equation}
    S_{\text{reas}} = [\mathcal{I}; x; a; r]
\end{equation}

Both strategies employ the same optimization process to ensure a fair comparison. The model is trained using the standard Causal Language Modeling (CLM) objective. Let $\theta$ denote the trainable parameters (LoRA adapters). We maximize the log-likelihood of the target label $y$ given the formatted input sequence $S$ (where $S$ is either $S_{\text{std}}$ or $S_{\text{reas}}$):
\begin{equation}
    \mathcal{L}(\theta) = - \sum_{t=1}^{T} \log P(y_t | y_{<t}, S; \theta)
\end{equation}
where $y_t$ is the $t$-th token of the target sentiment label. By comparing the performance of the model trained on $S_{\text{std}}$ versus $S_{\text{reas}}$, we aim to isolate the contribution of the reasoning component in the implicit sentiment analysis task.

\section{Experimental Setup}
\label{sec:experimental_setup}

\subsection{Datasets}
We evaluate our method on two standard benchmark datasets for ABSA: Restaurant and Laptop, originally derived from SemEval 2014 Task 4. In this work, we adopt the dataset configurations processed by Li et al. \cite{li-etal-2021-learning-implicit}, who explicitly partitioned the original data into distinct subsets of explicit and implicit sentiment to facilitate focused evaluation on ISA. The detailed statistics of these splits are presented in Table \ref{tab:datasets}.

\begin{table}[t]
\caption{Detailed statistics of the Restaurant and Laptop datasets, including sentiment label distribution and the proportion of implicit sentiment samples.}
\begin{center}
\begin{tabular}{lccccc}
\toprule
\textbf{Dataset} & \textbf{Positive} & \textbf{Neutral} & \textbf{Negative} & \textbf{Total} & \textbf{Implicit (\%)} \\
\midrule
\textbf{Restaurant} & 2892 & 1001 & 829 & 4722 & 27.47 \\
\hspace{3mm} Train & 2164 & 805 & 633 & 3602 & 28.59 \\
\hspace{3mm} Test & 728 & 196 & 196 & 1120 & 23.84 \\
\midrule
\textbf{Laptop} & 1328 & 994 & 629 & 2951 & 30.09 \\
\hspace{3mm} Train & 987 & 866 & 460 & 2313 & 30.87 \\
\hspace{3mm} Test & 341 & 128 & 169 & 638 & 27.27 \\
\bottomrule
\end{tabular}
\label{tab:datasets}
\end{center}
\end{table}

\subsection{Evaluation Metrics}
Following established conventions in ISA research \cite{10869316, 11214474}, we employ Accuracy as the primary evaluation metric. To provide a comprehensive assessment, we report:
\begin{itemize}
    \item Accuracy (All): Performance on the entire test set, encompassing both explicit and implicit samples.
    \item Accuracy (Implicit): Performance exclusively on the subset of implicit sentiment samples, which is the core focus of this study.
\end{itemize}

\subsection{Baselines}
To benchmark the effectiveness of our proposed method, we compare it against two categories of approaches:

\subsubsection{Prompt-based LLMs}
We employ three state-of-the-art (SOTA) general-purpose LLMs: GPT-4o, GPT-4o Mini, and Gemini 2.0 Flash. We design three prompting strategies to evaluate their zero-shot, few-shot and reasoning capabilities (see our github repository for the detailed prompts):
\begin{itemize}
    \item Zero-shot: Direct classification without examples.
    \item Few-shot: We provide 3 in-context examples (one for each polarity) randomly sampled from the training set of each dataset.
    \item Chain-of-Thought (CoT): We prompt the model to reason step-by-step before predicting the label.
\end{itemize}

\subsubsection{Fine-tuned Baselines}
We compare our results with existing supervised methods reported in the literature. For the majority of baselines, we directly cite the performance metrics from \cite{11214474} without re-implementation, while the results for the RVISA method are adopted from \cite{10869316}.
\begin{itemize}
    \item BERT-based: This category includes standard fine-tuning (BERT-SPC, BERT-ADA) and graph-based/syntax-aware models such as RGAT, BERT+CEPT, C$^3$DA, and BERT+SCAPT. Most of these methods are built upon BERT-Base (110M), while RGAT is also evaluated on the larger BERT-Large (340M) backbone.
    \item T5-based: We include the ABSA-ESA framework implemented on the T5-Base (220M) architecture.
    \item Flan-T5-based: We cover strong baselines including direct fine-tune, THOR, MT-ISA, and RVISA. To provide a comprehensive comparison across model scales, we report results for both the efficient Flan-T5-Base (250M) versions and the SOTA configurations leveraging the massive Flan-T5-XXL (11B) model.
\end{itemize}

\subsection{Implementation Details}

\subsubsection{Model Selection}
We utilize Gemma-7B (base version) as the backbone for our experiments. This model was selected due to its widespread adoption in the research community as a robust baseline for fine-tuning tasks. Furthermore, its 7B parameter size is well-suited to our available computational resources, offering a practical balance between performance and training efficiency compared to larger counterparts.

\subsubsection{Hyperparameters}
The model is fine-tuned using LoRA on a single NVIDIA A100 GPU (via Google Colab Pro). The training configuration is consistent for both Standard and Reasoning-Augmented settings:
\begin{itemize}
    \item LoRA Config: Rank $r=64$, Alpha $\alpha=128$, Dropout $=0.1$.
    \item Optimization: AdamW optimizer with a learning rate of $2e-4$ and a cosine learning rate scheduler.
    \item Batch Size: We use a per-device batch size of 8 with 2 gradient accumulation steps, resulting in an effective batch size of 16.
    \item Training Duration: The models are trained for 4 epochs with a maximum sequence length of 1024 tokens.
\end{itemize}

\subsubsection{Instruction Design}
The instruction prompts used for fine-tuning are designed to differentiate the input expectations:
\begin{itemize}
    \item Standard: ``Identify the sentiment polarity (positive, negative, or neutral) of the specific aspect in the following sentence''.
    \item Reasoning-Augmented: ``Identify the sentiment polarity (positive, negative, or neutral) of the specific aspect in the following sentence, \textit{utilizing the provided expert analysis}''.
\end{itemize}

\subsubsection{Reasoning Generation Setup}
To distill reasoning capabilities into the smaller model, we employ GPT-4o as the teacher model. We construct a specialized prompting protocol that enforces the teacher to simulate an expert linguistic analyst. Unlike standard CoT which often allows free-form generation, our prompt imposes a structured three-step reasoning path: (1) Contextual Anchoring, (2) Evidence Extraction (Explicit and Implicit), and (3) Synthesis. This structure ensures that the generated rationales strictly adhere to the logic of ISA. Crucially, this reasoning protocol is applied uniformly across both the Restaurant and Laptop datasets. The detailed prompt used for this generation process is provided in our github repository.

\section{Results Analysis}
\label{sec:results}

\subsection{Main Results}
The main experimental results on the Restaurant and Laptop datasets are summarized in Table \ref{tab:main_results}. We compare our proposed Gemma-7B models (Standard LoRA and Reasoning-Augmented LoRA) against established baselines. The performance is evaluated based on Accuracy across the entire test set (All) and ISA subset (ISA).

\begin{table}[t]
\centering
\caption{Performance Comparison (Accuracy \%). The category marked with $*$ denotes experiments implemented in this work, while results for other baselines are directly adopted from existing literature.}
\label{tab:main_results}
\setlength{\tabcolsep}{3pt}
\resizebox{\columnwidth}{!}{
\begin{tabular}{llccccc}
\toprule
\multirow{2}{*}{\textbf{Category}} & \multirow{2}{*}{\textbf{Model}} & \multirow{2}{*}{\textbf{Method}} & \multicolumn{2}{c}{\textbf{Restaurant}} & \multicolumn{2}{c}{\textbf{Laptop}} \\
\cmidrule(lr){4-5} \cmidrule(lr){6-7}
 & & & \textbf{All} & \textbf{ISA} & \textbf{All} & \textbf{ISA} \\
 
\midrule

\multirow{9}{*}[-1.25ex]{\textbf{Prompt-based}$^*$} 
 & \multirow{3}{*}{GPT-4o} & Zero-shot & 88.21 & 71.16 & 82.60 & 72.00 \\
 & & Few-shot & 86.88 & 70.79 & 80.09 & 67.43 \\
 & & Chain-of-Thought & 86.88 & 68.91 & 81.98 & 72.00 \\
\cmidrule(lr){2-7}
 & \multirow{3}{*}{GPT-4o Mini} & Zero-shot & 86.88 & 70.04 & 82.13 & 71.43 \\
 & & Few-shot & 85.00 & 67.42 & 81.35 & 72.57 \\
 & & Chain-of-Thought & 86.16 & 66.67 & 81.35 & 70.86 \\
\cmidrule(lr){2-7}
 & \multirow{3}{*}{Gemini 2.0 Flash} & Zero-shot & 84.20 & 70.41 & 79.00 & 64.57 \\
 & & Few-shot & 87.76 & 69.66 & 81.97 & 67.43 \\
 & & Chain-of-Thought & 82.17 & 67.79 & 75.39 & 57.71 \\
 
\midrule

\multirow{7}{*}[-1ex]{\textbf{BERT-based}} 
 & \multirow{6}{*}{BERT-Base (110M)} & BERT-SPC & 83.57 & 65.54 & 78.22 & 69.54 \\
 & & BERT-ADA & 87.14 & 65.92 & 78.96 & 70.11 \\
 & & RGAT & 86.60 & 67.79 & 78.21 & 72.99 \\
 & & BERT$_{Asp}$ + CEPT & 87.50 & 67.79 & 81.66 & 75.86 \\
 & & C$^3$DA & 86.93 & 65.54 & 80.61 & 73.57 \\
 & & BERT$_{Asp}$ + SCAPT & 89.11 & 72.28 & 82.76 & 77.59 \\
\cmidrule(lr){2-7}
 & BERT-Large (340M) & RGAT & 86.87 & 68.05 & 80.25 & 75.43 \\
 
\midrule

\textbf{T5-based} & T5-Base (220M) & ABSA-ESA & 88.29 & 70.78 & 82.44 & 80.00 \\

\midrule

\multirow{10}{*}[-2.5ex]{\textbf{Flan-T5-based}} 
 & \multirow{5}{*}{Flan-T5-Base (250M)} & Direct fine-tune & 86.88 & 65.17 & 81.98 & 73.71 \\
 & & THOR & 87.68 & 68.54 & 81.66 & 74.29 \\
 & & MT-ISA$_I$ & 88.21 & 70.41 & 82.91 & 80.57 \\
 & & MT-ISA$_O$ & 85.89 & 66.29 & 81.98 & 75.43 \\
 & & RVISA$_v$ & 86.43 & 65.92 & 80.72 & 73.71 \\
 & & RVISA$_g$ & 86.61 & 66.67 & 81.19 & 75.43 \\
\cmidrule(lr){2-7}
 & \multirow{5}{*}{Flan-T5-XXL (11B)} & Direct fine-tune & 89.29 & 75.28 & 81.82 & 75.43 \\
 & & THOR & 88.57 & 73.03 & 82.29 & 76.57 \\
 & & MT-ISA$_I$ & 91.34 & 81.27 & 85.40 & 86.29 \\
 & & MT-ISA$_O$ & \textbf{92.68} & \textbf{84.27} & 85.74 & \textbf{88.57} \\
 & & RVISA$_v$ & 91.25 & 81.65 & 86.52 & 87.43 \\
 & & RVISA$_g$ & 91.52 & 82.02 & \textbf{86.68} & 88.00 \\
\midrule

\multirow{2}{*}{\textbf{Ours}} & \multirow{2}{*}{Gemma-7B} & Reasoning LoRA & \underline{89.63} & 74.53 & 83.70 & 76.00 \\
 & & Standard LoRA & 89.46 & \underline{75.28} & \underline{84.01} & \underline{81.14} \\
\bottomrule
\end{tabular}
}
\end{table}

\paragraph{Superiority over baselines} 
As observed in Table \ref{tab:main_results}, our fine-tuned Gemma-7B models (both Standard and Reasoning-Augmented) consistently outperform all Prompt-based approaches (GPT-4o, GPT-4o Mini, Gemini 2.0 Flash) and the majority of supervised baselines based on BERT, T5-Base, and Flan-T5-Base. Notably, in the Restaurant domain, our Reasoning-Augmented model achieves 89.63\% accuracy, surpassing the strong BERT$_{Asp}$ + SCAPT (89.11\%) and Flan-T5-Base variants. Meanwhile, in the Laptop domain, our Standard LoRA model demonstrates exceptional robustness, achieving an ISA accuracy of 81.14\%, which significantly outperforms T5-Base (80.00\%), Flan-T5-Base and BERT-based variants. This demonstrates that parameter-efficient fine-tuning (LoRA) on a 7B backbone effectively captures implicit sentiment nuances better than smaller fully fine-tuned models or zero-shot prompts on larger models.

\paragraph{Competitive Performance against SOTA} 
While the SOTA methods (MT-ISA and RVISA) using Flan-T5-XXL (11B) achieve higher scores, it is crucial to highlight the efficiency trade-off. Our method utilizes a 7B model optimized via LoRA (updating $<1\%$ parameters), whereas the SOTA baselines rely on an 11B parameter model, which requires significantly higher computational resources for training and inference. Despite this disadvantage, our Gemma-7B models remain highly competitive, narrowing the performance gap significantly compared to the Flan-T5-Base (250M) baselines.

\subsection{Impact of Reasoning Augmentation}
A key objective of this study was to investigate whether augmenting input with explicit reasoning traces enhances performance. The results present a nuanced picture, revealing a strong domain dependency.

\paragraph{Benefit in Subjective Domains}
In the Restaurant dataset, the Reasoning-Augmented LoRA achieves a slight performance gain over the Standard LoRA (89.63\% vs. 89.46\%). Restaurant reviews are predominantly characterized by subjective experiences, where sentiment is often embedded in descriptions of service, atmosphere, or culinary nuance (e.g., ``small portions'' implying negative sentiment). In such contexts, the generated reasoning traces act as a semantic bridge, helping the model connect explicit textual descriptions with implicit pragmatic implications that might otherwise be missed by a standard shallow analysis.

\paragraph{Challenges in Objective/Technical Domains} 
Conversely, in the Laptop dataset, we observe a notable performance degradation with Reasoning Augmentation, particularly in the ISA subset where accuracy drops from 81.14\% to 76.00\%. We attribute this phenomenon to the objective nature of the domain. Laptop reviews frequently hinge on technical specifications (e.g., ``8GB RAM'', ``plastic casing'') where sentiment is derived from a strict deviation from functional norms rather than linguistic expression. In this scenario, the reasoning generated by the teacher model (GPT-4o) may introduce information noise or superfluous elaborations. Instead of aiding classification, these lengthy rationales can cause attention dilution, distracting the smaller 7B model from the concise technical cues that are decisive for the label. Furthermore, the teacher model might attempt to over-interpret neutral specifications using linguistic logic, leading to misaligned inferences that confuse the student model. This finding suggests that while reasoning is valuable for subjective interpretation, its integration in objective domains requires careful filtering to avoid overwhelming the model with irrelevant context.

\subsection{Qualitative Analysis and Case Study}
To investigate the underlying causes of the performance variations, we conducted a qualitative analysis of the misclassified instances, focusing on the discrepancy between the Standard and Reasoning-Augmented models. Fig. \ref{fig:case_study} illustrates two representative examples that highlight the domain-dependent behavior of our framework.

\begin{figure}[t]
    \centering
    \fbox{\parbox{0.45\textwidth}{
        \small
        \textbf{Example 1: Restaurant (Reasoning Wins)} \\
        \textit{Input:} ``Entrees include classics like lasagna, fettuccine Alfredo and chicken parmigiana''. \\
        \textit{Aspect:} chicken parmigiana \quad \textit{Gold Label:} \textbf{Neutral} \\
        \textit{Reasoning (Condensed):} The aspect is mentioned within the context of a \textbf{list of well-known Italian dishes}... The nature of the statement is \textbf{objective}, listing menu items without direct opinion words...\\
        \textit{Standard Prediction:} Positive $\times$ \\
        \textit{Reasoning Prediction:} \textbf{Neutral} $\checkmark$
        
        \rule{\linewidth}{0.5pt}
        
        \textbf{Example 2: Laptop (Reasoning Fails)} \\
        \textit{Input:} ``No installation disk (DVD) is included''. \\
        \textit{Aspect:} installation disk (DVD) \quad \textit{Gold Label:} \textbf{Neutral} \\
        \textit{Reasoning (Condensed):} The nature of the statement is an \textbf{objective fact}, explicitly noting the absence of the disk... However, \textbf{implicit sentiment} can be inferred... the absence likely \textbf{creates inconvenience}... implicitly suggests a \textbf{negative sentiment} as it detracts from user utility... \\
        \textit{Standard Prediction:} \textbf{Neutral} $\checkmark$ \\
        \textit{Reasoning Prediction:} Negative $\times$
    }}
    \caption{Qualitative comparison between Standard and Reasoning-Augmented models. Reasoning traces are condensed to highlight the conflict between objective framing and implicit inference.}
    \label{fig:case_study}
\end{figure}

\paragraph{Benefit of Reasoning in Subjective Contexts}
Example 1 (Restaurant) highlights a common failure mode of the Standard model: relying on shallow lexical shortcuts. The presence of the word ``classics'' likely triggered a \textit{Positive} prediction, disregarding the syntactic structure of the sentence. In contrast, the Reasoning-Augmented model correctly predicts \textit{Neutral}. Crucially, although the teacher model's rationale wanders into implying positive satisfaction, it explicitly anchors the text as ``listing menu items'' and ``objective''. Our fine-tuned model successfully attends to this contextual grounding, recognizing that a menu description—despite containing positive entities—is fundamentally a neutral enumeration.

\paragraph{The ``Over-interpretation'' Error Pattern}
In contrast, the Laptop domain reveals a systematic error pattern in the Reasoning-Augmented model. As depicted in Example 2, the model misclassifies a neutral factual statement (``only 2 USB ports'') as Negative. Error Analysis on the Laptop ISA subset indicates that this is not an isolated incident; the Reasoning-Augmented model misclassifies ``Neutral'' technical descriptions as ``Negative'' or ``Positive'' significantly more often than the Standard baseline. We attribute this to an \textit{over-interpretation} phenomenon: the reasoning generation process, trained to mimic expert analysts, tends to hallucinate pragmatic implications from strict technical specifications (e.g., treating a standard feature as a defect). Consequently, the model assigns sentiment based on these hallucinated inconveniences rather than the objective text, leading to lower accuracy in technical domains.

\section{Conclusion and Future Work}
\label{sec:conclusion}
We presented a parameter-efficient framework for Implicit Sentiment Analysis using Gemma-7B fine-tuned with LoRA and distilled reasoning traces. Our experiments demonstrate that this approach significantly outperforms BERT-based and prompt-based baselines, achieving parity with state-of-the-art 11B-parameter models. Crucially, our analysis reveals that the efficacy of reasoning augmentation is \textit{domain-dependent}: while beneficial for bridging semantic gaps in subjective reviews (Restaurant), explicit reasoning can introduce information noise in objective technical contexts (Laptop). Future work will focus on developing \textit{adaptive reasoning filtering} mechanisms to mitigate such noise and investigating the framework's robustness through cross-domain generalization.

\section*{Acknowledgement}
This research is funded by Vietnam National University, Ho Chi Minh City (VNU-HCM) under grant number C2026-26-10.

\bibliographystyle{IEEEtran}
\bibliography{reference}
\end{document}